\documentclass[10pt,twocolumn,letterpaper]{article}

\usepackage{cvpr}              

\usepackage{amsmath,amssymb,amsfonts}
\usepackage{algorithmic}
\usepackage{algorithm}
\usepackage{graphicx}
\usepackage{booktabs}
\usepackage{multirow}
\usepackage{xspace}
\usepackage{microtype}
\usepackage{subcaption}
\usepackage{tikz}
\usetikzlibrary{calc}

\newcommand{\bagsac}{BA-GSAC\xspace}
\newcommand{\gsac}{GSAC\xspace}

\DeclareMathOperator{\E}{\mathbb{E}}

\usepackage[accsupp]{axessibility} 

\definecolor{cvprblue}{rgb}{0.21,0.49,0.74}
\usepackage[pagebackref,breaklinks,colorlinks,allcolors=cvprblue]{hyperref}

\def\paperID{WAD-6}
\def\confName{CVPR}
\def\confYear{2026}

\title{When Does Adaptive Guidance Help? Belief-Aware Privileged\\Distillation for Autonomous Driving Under Partial Observability}

\author{Mehmet Haklidir\\
TUBITAK BILGEM Artificial Intelligence Institute, Turkey\\
{\tt\small mehmet.haklidir@tubitak.gov.tr}}

\begin{document}
\maketitle
\begin{abstract}
Guided Soft Actor-Critic (GSAC) distills knowledge from a privileged full-state teacher to a partial-observation student for autonomous driving, but uses a fixed distillation coefficient~$\lambda$ regardless of the agent's uncertainty.
We present \emph{Belief-Aware GSAC} (\bagsac), which modulates $\lambda$ via ensemble disagreement, and use it as a testbed for a systematic empirical study asking: \emph{when does adaptive guidance actually help?}
Evaluating five strategies (fixed $\lambda \in \{0.01, 0.1\}$, adaptive, linear decay, and vanilla SAC) across three POMDP difficulty levels on Highway-Env, we find that preliminary single-seed runs suggest benefits under mild and moderate partial observability, but under severe occlusion (evaluated with 3 seeds for all methods) the adaptive coefficient collapses to $\lambda_{\min}$ within $\sim$3K steps.
We trace this to an \emph{observability blindness} phenomenon: because the ensemble predicts \emph{partial observations}, it achieves low disagreement even under heavy occlusion, modeling what is visible but unable to detect what is \emph{missing}.
We diagnose the root cause and propose an architectural fix (training the ensemble on full-state predictions using the guiding actor's privileged access); while not validated here, we show that even with current limitations, the warmup phase provides measurable stabilization (CV=13.3\% vs.\ 29.8\% for constant $\lambda\!=\!0.01$).
In fact, a simple deterministic linear decay schedule achieves the best severe-POMDP performance across all metrics (mean 116.5, CV=8.9\%), suggesting that the scheduling effect, not the ensemble, drives the stability benefit.
These findings provide practical guidance for designing uncertainty-aware teacher-student frameworks and highlight ensemble prediction targets as an important design choice.
\end{abstract}

\section{Introduction}
\label{sec:intro}

Autonomous vehicles must navigate environments where the true state is never fully observable.
Sensor occlusion, adverse weather, and limited field of view create a partially observable Markov decision process (POMDP), where the agent must infer hidden aspects of the world from noisy, incomplete observations~\cite{kaelbling1998planning}.
This partial observability is not merely a theoretical concern; it directly impacts safety-critical decisions such as lane changes, intersection navigation, and highway merging.

Reinforcement learning (RL) offers a principled framework for sequential decision-making in autonomous driving~\cite{kiran2021deep}.
Among RL methods, Soft Actor-Critic (SAC)~\cite{haarnoja2018soft} has demonstrated strong performance across continuous control tasks due to its entropy-regularized policy optimization and sample efficiency.
However, standard SAC assumes full state observability, and its performance degrades significantly in POMDP settings where critical state dimensions are hidden.

To address this, Guided SAC (\gsac)~\cite{haklidir2021guided} introduces a teacher-student framework where a \emph{guiding actor} with full state access distills knowledge to a \emph{control actor} operating on partial observations through a dedicated distillation network.

A critical limitation of \gsac is its use of a \emph{fixed distillation coefficient} throughout training.
This creates two failure modes: (1)~in states where the agent already has a good model of the environment, fixed guidance introduces unnecessary bias that can prevent the control actor from developing its own effective strategies; (2)~in novel or highly uncertain states, the fixed coefficient may provide insufficient guidance when the agent needs it most.
The optimal level of guidance should be \emph{state-dependent and uncertainty-aware}.

\begin{figure*}[t]
    \centering
    \resizebox{0.95\textwidth}{!}{
    \begin{tikzpicture}[
        comp/.style={rectangle, draw, rounded corners=4pt, minimum height=1.1cm, align=center, font=\normalsize, line width=0.9pt},
        envbox/.style={comp, fill=gray!8, draw=gray!50, minimum width=2.8cm},
        bluebox/.style={comp, fill=blue!8, draw=blue!55, minimum width=2.8cm},
        orangebox/.style={comp, fill=orange!8, draw=orange!55, minimum width=2.8cm},
        greenbox/.style={comp, fill=green!8, draw=green!50, minimum width=2.8cm},
        redbox/.style={comp, fill=red!6, draw=red!50, minimum width=2.8cm},
        redhigh/.style={comp, fill=red!15, draw=red!65, minimum width=2.8cm, line width=1.2pt},
        arr/.style={->, >=stealth, line width=1pt},
        bluearr/.style={arr, blue!60},
        orangearr/.style={arr, orange!60},
        redarr/.style={arr, red!55, dashed, line width=1.1pt},
        lbl/.style={font=\small, text=gray!40, inner sep=2pt},
        hlbl/.style={font=\small\bfseries, text=red!60},
    ]

    \node[envbox] (env) at (0, 0) {Highway-Env\\(POMDP)};

    \node[bluebox] (fullst) at (3.8, 1.5) {Full state $s_t$};
    \node[orangebox] (partobs) at (3.8, -1.5) {Partial obs $o_t$};

    \draw[arr, gray!60] (env.east) -- ++(1.0,0) coordinate (fork) |- (fullst.west);
    \draw[arr, gray!60] (fork) |- (partobs.west);
    \node[lbl, right] at (1.6, 0.7) {privileged};
    \node[lbl, right] at (1.6, -0.7) {sensor};

    \node[bluebox] (guiding) at (8.2, 1.5) {Guiding actor\\$\pi_g(a|s_t)$};
    \node[orangebox] (control) at (8.2, -1.5) {Control actor\\$\pi_c(a|h_t)$};

    \draw[bluearr] (fullst) -- (guiding);
    \draw[orangearr] (partobs) -- node[lbl, below] {history $h_t$} (control);

    \node[greenbox] (critic) at (12.5, 0) {Twin Q-network\\$Q(s_t, a)$};

    \draw[bluearr] (guiding.east) -- ++(0.5,0) |- (critic.west);
    \draw[orangearr] (control.east) -- ++(0.5,0) |- (critic.west);

    \draw[redarr] (guiding.south) -- node[hlbl, right, align=left] {distill\\$\lambda_t \cdot \|\pi_c - D\|^2$} (control.north);

    \node[redbox] (ensemble) at (3.2, -3.8) {Ensemble $\{f_{\phi_i}\}_{i=1}^N$\\predict $\hat{o}_{t+1}$};

    \node[redhigh] (adaptive) at (9.0, -3.8) {\textbf{Adaptive $\lambda_t$}\\$\lambda_t = g(u_t)$};

    \draw[orangearr] (partobs.south) -- node[lbl, right] {$h_t, a_t$} (ensemble.north);
    \draw[arr, red!55, line width=1.1pt] (ensemble) -- node[lbl, above, yshift=2pt] {disagreement $u_t$} (adaptive);
    \draw[redarr, line width=1.3pt] (adaptive.north) -- node[hlbl, right] {modulates} (control.south);

    \draw[gray!30, dashed, line width=0.6pt] (-1.2, -2.7) -- (14, -2.7);
    \node[font=\small\itshape, text=gray!45, anchor=west] at (-1.2, -2.45) {GSAC base (above)};
    \node[font=\small\itshape, text=red!50, anchor=west] at (-1.2, -2.95) {BA-GSAC contribution (below)};

    \node[font=\scriptsize, text=gray!50, anchor=west, align=left] at (11, -2.0) {CTDE: critic uses $s_t$\\at training; only $\pi_c(h_t)$\\deployed at test time};

    \end{tikzpicture}
    }
    \caption{\textbf{BA-GSAC architecture.} The environment provides full state to the guiding actor (blue, top) and partial observations to the control actor (orange, middle). GSAC distills the guiding actor's knowledge to the control actor via a $\lambda$-weighted loss. \textbf{Our contribution} (red, below dashed line): an ensemble of forward models measures disagreement $u_t$, which adaptively modulates $\lambda_t$ so that guidance is strong when the agent is uncertain and weak when confident.}
    \label{fig:overview}
\end{figure*}
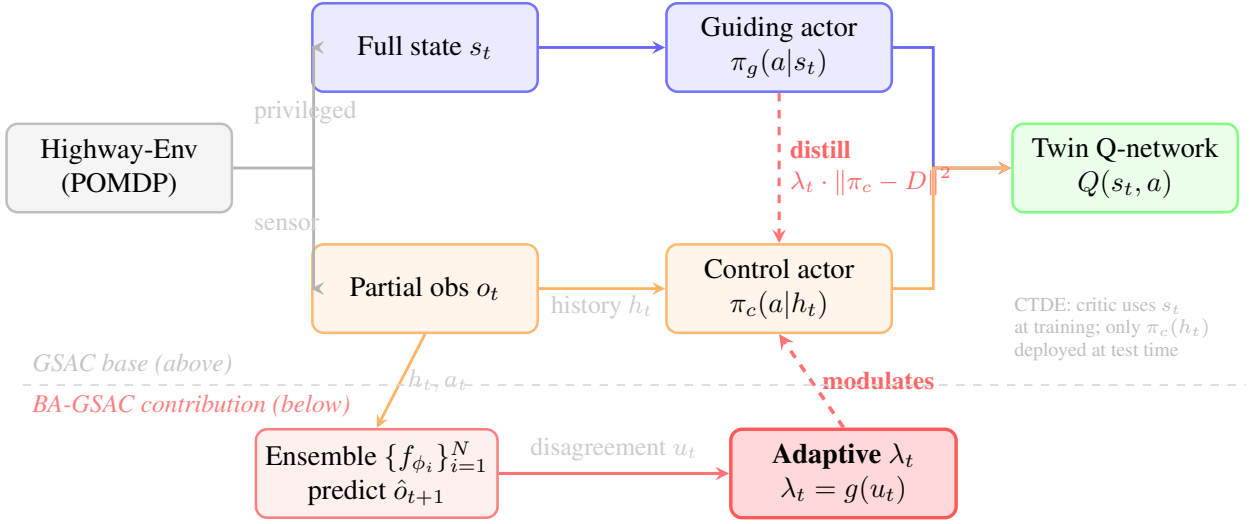

We propose \emph{Belief-Aware Guided SAC} (\bagsac), which addresses this limitation by dynamically modulating the distillation coefficient based on epistemic uncertainty (Fig.~\ref{fig:overview}).
The central idea is that an ensemble of forward dynamics models can serve as a proxy for the agent's epistemic uncertainty: when ensemble members disagree about state transitions, the agent is in an unfamiliar region of the state space and should rely more heavily on the privileged guiding actor.
Conversely, when the ensemble agrees, the agent has sufficient knowledge to act autonomously.

However, our systematic evaluation reveals that the answer to ``when does adaptive guidance help?'' is mixed.
Through controlled experiments across three POMDP difficulty levels with a fixed $\lambda\!=\!0.01$ baseline (matching BA-GSAC's asymptotic behavior), we find that adaptive guidance provides clear benefits under mild and moderate POMDP, but collapses under severe conditions.
We trace this to an architectural issue we term \emph{observability blindness}: because the ensemble predicts partial observations, it cannot detect uncertainty from \emph{missing} information.
This diagnosis points to a plausible fix (training the ensemble on full-state predictions), which we leave for future work.

Our contributions are:
\begin{enumerate}
    \item We integrate ensemble disagreement with GSAC's distillation coefficient and conduct a systematic evaluation asking \emph{when} adaptive guidance helps across three POMDP severity levels.
    \item We empirically demonstrate that observation-space ensemble prediction creates a structural blind spot under occlusion (\emph{observability blindness}): the ensemble cannot detect uncertainty from masked state dimensions. This finding establishes the prediction target as a critical design choice for ensemble-based uncertainty in POMDPs, with implications beyond our specific method.
    \item We demonstrate through a fixed $\lambda\!=\!0.01$ baseline that BA-GSAC's advantage is not simply ``less guidance'' but comes from the warmup phase's adaptive scheduling, with measurably lower cross-seed variance (CV=13.3\% vs.\ 29.8\%).
    \item We discuss implications for world model-based simulators~\cite{bruce2024genie} and propose full-state ensemble prediction as a plausible architectural remedy (not validated here).
\end{enumerate}

\section{Related Work}
\label{sec:related}

\paragraph{Reinforcement Learning for Autonomous Driving.}
Deep RL has been widely applied to autonomous driving tasks including lane keeping~\cite{sallab2017deep}, highway decision-making~\cite{leurent2019approximate}, and intersection navigation~\cite{isele2018navigating}.
Model-free approaches such as DQN~\cite{mnih2015human} and SAC~\cite{haarnoja2018soft} have shown promise, but typically assume full state observability.
Recent work has explored recurrent architectures~\cite{hausknecht2015deep} and transformer-based memory~\cite{chen2021decision} to handle partial observability.
Ni~\etal~\cite{ni2022recurrent} showed that recurrent model-free RL can be a strong baseline for many POMDPs, often matching or exceeding more complex approaches, which is an important reference point for our observation history approach.
MetaDrive~\cite{li2022metadrive} provides a more scalable driving RL benchmark with diverse scenarios; extending our evaluation to such environments is an important direction.

\paragraph{Teacher-Student and Privileged Learning.}
The idea of learning from a privileged expert has deep roots in autonomous driving.
Chen~\etal~\cite{chen2020learning} demonstrated that imitation learning from a privileged agent with ground-truth perception significantly outperforms learning from raw sensors alone.
Zhang~\etal~\cite{zhang2021roach} extended this to RL coaching in CARLA, where a privileged RL agent supervises an imitation learner for end-to-end urban driving.
Pinto~\etal~\cite{pinto2018asymmetric} proposed asymmetric actor-critic, using privileged information only in the critic during training.
Baisero and Amato~\cite{baisero2022unbiased} formalized the asymmetric information setting, showing that unbiased value estimation requires careful handling of the information gap.
Guided SAC~\cite{haklidir2021guided} further extends this approach to a dual-actor RL framework with explicit knowledge distillation.
However, all these methods use fixed-strength knowledge transfer that does not adapt to the agent's evolving competence.
Czarnecki~\etal~\cite{czarnecki2019distilling} analyzed when and how policy distillation helps, finding that the benefit depends on the capacity and training stage of the student; our work extends this analysis to the POMDP setting with adaptive scheduling.

\paragraph{Uncertainty Estimation in RL.}
Epistemic uncertainty quantification has been used in RL for exploration~\cite{osband2016deep,pathak2017curiosity,burda2019exploration}, safe control~\cite{berkenkamp2017safe}, and model-based planning~\cite{chua2018deep,sekar2020planning}.
Ensemble methods are particularly attractive due to their simplicity~\cite{lakshminarayanan2017simple}; Lee~\etal~\cite{lee2020sunrise} proposed SUNRISE, which uses ensemble-based weighted Bellman backups for more stable RL training. Alternatives include Bayesian neural networks and latent-variable world models such as DreamerV3~\cite{hafner2023mastering}, which handle partial observability via learned latent states (RSSM).
We choose ensembles for their simplicity and compatibility with the GSAC framework, but note that latent-state models may offer richer uncertainty representations.
Our work uniquely combines ensemble uncertainty with the teacher-student framework, and our observability blindness finding (Sec.~\ref{sec:guidance_dynamics}) characterizes and empirically demonstrates a failure mode of observation-space ensemble prediction under occlusion.
While ensemble prediction targets have been studied for exploration~\cite{pathak2017curiosity} and value estimation~\cite{lee2020sunrise}, the interaction between prediction target space and partial observability has not, to our knowledge, been analyzed: prior ensemble RL work assumes full-state or learned-latent targets, making the failure mode under raw masked observations easy to overlook in practice.

\paragraph{World Models and Simulation for Driving.}
Generative world models~\cite{ha2018world,hafner2023mastering,bruce2024genie} produce inherently partially observable environments, motivating POMDP-robust policies.
Think2Drive~\cite{li2024think2drive} uses latent world models for driving, further motivating uncertainty-aware policy architectures.

\section{Method}
\label{sec:method}

We first formalize the POMDP setting for autonomous driving (Sec.~\ref{sec:pomdp}), review Guided SAC (Sec.~\ref{sec:gsac_review}), then present our Belief-Aware extension (Sec.~\ref{sec:bagsac}).

\subsection{POMDP Formulation for Driving}
\label{sec:pomdp}

We model the autonomous driving task as a POMDP $(\mathcal{S}, \mathcal{A}, \mathcal{O}, T, O, R, \gamma)$, where $\mathcal{S}$ is the full state space (positions, velocities of all vehicles), $\mathcal{A}$ is the continuous action space (acceleration, steering), $\mathcal{O}$ is the observation space, $T(s'|s,a)$ is the transition function, $O(o|s)$ is the observation function, $R(s,a)$ is the reward, and $\gamma$ is the discount factor.

In practice, the ego vehicle observes a noisy, occluded subset of the full state:
\begin{equation}
    o_t = O(s_t) = M_t \odot s_t + \epsilon_t,
    \label{eq:observation}
\end{equation}
where $M_t \in \{0,1\}^{|\mathcal{S}|}$ is a stochastic occlusion mask (vehicles may disappear from sensor range) and $\epsilon_t \sim \mathcal{N}(0, \sigma^2 I)$ represents sensor noise.

\subsection{Guided SAC Review}
\label{sec:gsac_review}

Guided SAC~\cite{haklidir2021guided} employs two actors operating in alternating interaction with the environment:

\textbf{Guiding actor} $\pi_g(a|s;\theta_g)$: Has access to the full state $s$ and optimizes the standard maximum-entropy objective:
\begin{equation}
    J_g = \E_{s,a \sim \pi_g} \left[ Q(s,a) - \alpha \log \pi_g(a|s) \right].
    \label{eq:guiding_obj}
\end{equation}

\textbf{Control actor} $\pi_c(a|h_t;\theta_c)$: Operates on observation history $h_t = (o_{t-K}, \ldots, o_t)$ and is trained with both the RL objective and a distillation term:
\begin{equation}
    J_c = \E_{h_t,a \sim \pi_c} \left[ Q(s,a) - \alpha \log \pi_c(a|h_t) \right] - \lambda \| \pi_c(h_t) - D(h_t) \|^2,
    \label{eq:control_obj}
\end{equation}
where $D(h)$ is a distillation DNN that maps observation histories to the guiding actor's actions, and $\lambda$ is the \emph{fixed} distillation coefficient.

Both actors share a twin Q-network $Q(s,a)$ taking full state, following centralized-training-decentralized-execution (CTDE): during training, both critic and guiding actor receive full state $s$; at test time, only $\pi_c(a|h_t)$ is deployed with no privileged access.

\subsection{Belief-Aware Guided SAC}
\label{sec:bagsac}

The key limitation of \gsac is that $\lambda$ remains constant throughout training and across states.
We propose to make $\lambda$ \emph{adaptive}, increasing guidance when the agent is uncertain and decreasing it when confident.

\subsubsection{Epistemic Uncertainty via Ensemble Disagreement}
\label{sec:ensemble}

We maintain an ensemble of $N$ forward dynamics models $\{f_{\phi_i}\}_{i=1}^N$, each a 2-layer MLP (64 hidden units) trained independently with MSE loss to predict the next \emph{partial} observation given the current observation history and action:
\begin{equation}
    \hat{o}_{t+1}^{(i)} = f_{\phi_i}(h_t, a_t), \quad i = 1, \ldots, N.
    \label{eq:ensemble_pred}
\end{equation}
Each ensemble member is initialized with a different random seed and trained on uniformly sampled replay buffer minibatches (no bootstrap sampling; random initialization alone can provide sufficient diversity for disagreement-based uncertainty~\cite{lakshminarayanan2017simple}, though bootstrap resampling could further strengthen epistemic signals).
Inputs and targets are raw (unnormalized) kinematics features to preserve the zero structure of occluded dimensions; this means disagreement magnitude can be dominated by higher-scale features (e.g., velocities vs.\ positions), a limitation that feature normalization or latent-space prediction could address.
Occluded vehicle features are zero in both the input history $h_t$ and the target $o_{t+1}$, which is critical to the observability blindness phenomenon analyzed in Sec.~\ref{sec:guidance_dynamics}.

The epistemic uncertainty is quantified by the pairwise disagreement among ensemble members:
\begin{equation}
    u(h_t, a_t) = \frac{2}{N(N-1)} \sum_{i < j} \| \hat{o}_{t+1}^{(i)} - \hat{o}_{t+1}^{(j)} \|^2.
    \label{eq:disagreement}
\end{equation}

This disagreement captures \emph{epistemic} (knowledge) uncertainty: it is high in regions of the state space where the agent has limited experience and low where the dynamics are well-understood.
It decreases as the ensemble accumulates more data, distinguishing it from aleatoric uncertainty which is irreducible.

\subsubsection{Adaptive Guidance Modulation}
\label{sec:adaptive}

We replace the fixed $\lambda$ with an adaptive coefficient $\lambda_t$ that is a function of the current epistemic uncertainty:
\begin{equation}
    \lambda_t = \lambda_{\min} + (\lambda_{\max} - \lambda_{\min}) \cdot \sigma\left(\frac{u_t - u_{\text{lo}}}{u_{\text{hi}} - u_{\text{lo}}}\right),
    \label{eq:adaptive_lambda}
\end{equation}
where $\sigma(\cdot) = \text{clip}(\cdot, 0, 1)$ is a clipped linear mapping, and $u_{\text{lo}}, u_{\text{hi}}$ are reference uncertainty bounds calibrated during a warmup phase.

\textbf{Warmup calibration.}
During the first $W$ training steps, we collect uncertainty values and set $u_{\text{lo}} = P_{10}(\{u_t\}_{t=1}^W)$ and $u_{\text{hi}} = P_{90}(\{u_t\}_{t=1}^W)$ as the 10th and 90th percentiles.
This data-driven calibration avoids environment-specific hyperparameter tuning.

The resulting \bagsac control actor objective becomes:
\begin{equation}
    J_c^{\text{BA}} = \E \left[ Q(s,a) - \alpha \log \pi_c(a|h_t) \right] - \lambda_t \| \pi_c(h_t) - D(h_t) \|^2.
    \label{eq:bagsac_obj}
\end{equation}

The full update procedure is given in Algorithm~\ref{alg:bagsac}.

\begin{algorithm}[t]
\caption{Belief-Aware Guided SAC (\bagsac)}
\label{alg:bagsac}
\begin{algorithmic}[1]
\REQUIRE Guiding actor $\pi_g$, control actor $\pi_c$, critic $Q$, ensemble $\{f_{\phi_i}\}_{i=1}^N$, distillation DNN $D$
\STATE Initialize replay buffer $\mathcal{D}$, warmup buffer $\mathcal{U}$
\FOR{each training step $t$}
    \STATE // \textit{Iterative interaction (alternating actors)}
    \IF{$t$ is even}
        \STATE $a_t \sim \pi_g(\cdot | s_t)$ \hfill $\triangleright$ Guiding actor (full state)
    \ELSE
        \STATE $a_t \sim \pi_c(\cdot | h_t)$ \hfill $\triangleright$ Control actor (history)
    \ENDIF
    \STATE Execute $a_t$, observe $o_{t+1}, r_t$; store in $\mathcal{D}$
    \STATE // \textit{Compute epistemic uncertainty}
    \STATE $u_t \leftarrow$ EnsembleDisagreement$(h_t, a_t)$ \hfill [Eq.~\ref{eq:disagreement}]
    \STATE $\lambda_t \leftarrow$ AdaptiveGuidance$(u_t)$ \hfill [Eq.~\ref{eq:adaptive_lambda}]
    \STATE // \textit{Gradient updates}
    \STATE Update $Q$ via TD targets from $\pi_g$ \hfill [Eq.~\ref{eq:guiding_obj}]
    \STATE Update $\pi_g$ via maximum entropy \hfill [Eq.~\ref{eq:guiding_obj}]
    \STATE Update $\pi_c$ with adaptive distillation \hfill [Eq.~\ref{eq:bagsac_obj}]
    \STATE Update $D$ via supervised loss on $\pi_g$ actions
    \STATE Update ensemble $\{f_{\phi_i}\}$ on transition data
    \STATE Update target networks via Polyak averaging
\ENDFOR
\end{algorithmic}
\end{algorithm}

\subsubsection{Interpretation as Epistemic Uncertainty Proxy}

Ensemble disagreement provides a heuristic proxy for epistemic uncertainty: in regions with sparse training data, ensemble members disagree, while in well-visited regions they converge~\cite{lakshminarayanan2017simple}.
In a POMDP, this can loosely be related to belief dispersion: high disagreement suggests the agent's implicit belief over hidden states is diffuse, while low disagreement suggests a concentrated belief.

However, we note an important caveat: because the ensemble predicts \emph{partial observations} $\hat{o}_{t+1}$, its disagreement reflects uncertainty about \emph{observable} dynamics.
Uncertainty arising from \emph{unobserved} state dimensions (occluded vehicles) may not produce disagreement if the observable dynamics are simple.
We analyze the consequences of this architectural choice in Sec.~\ref{sec:guidance_dynamics} and discuss a remedy in Sec.~\ref{sec:conclusion}.

\section{Experiments}
\label{sec:experiments}

We evaluate \bagsac against fixed \gsac and vanilla SAC on highway driving scenarios under varying levels of partial observability.

\subsection{Experimental Setup}
\label{sec:setup}

\paragraph{Environment.}
We use Highway-Env~\cite{leurent2018environment} (v1.8.2), a lightweight driving simulator built on the Gymnasium~\cite{towers2024gymnasium} interface.
We choose this environment as a controlled testbed where we can isolate the effect of occlusion rate on adaptive guidance, without confounds from visual rendering or complex sensor models; extension to richer simulators~\cite{li2022metadrive} is left for future work.
We evaluate on the \texttt{highway-v0} scenario with \texttt{ContinuousAction} type: a multi-lane highway with dense traffic where the ego vehicle must navigate at high speed while avoiding collisions.
The observation space consists of kinematics features (presence, $x$, $y$, $v_x$, $v_y$) for the ego vehicle and its 4 nearest neighbors, yielding a $5 \times 5$ matrix (flattened to 25 dimensions).
Actions are continuous 2D: longitudinal acceleration $\in [-1, 1]$ and lateral steering $\in [-1, 1]$, mapped to physical acceleration and heading change by the simulator.

\paragraph{POMDP Construction.}
We create three levels of partial observability by applying observation noise and stochastic vehicle occlusion to the full state (Table~\ref{tab:pomdp_levels}):

\begin{table}[t]
\centering
\caption{POMDP levels used in experiments.}
\label{tab:pomdp_levels}
\small
\begin{tabular}{lccl}
\toprule
\textbf{Level} & \textbf{Noise $\sigma$} & \textbf{Occl. \%} & \textbf{Driving Analogy} \\
\midrule
Mild     & 0.02 & 10\% & Light rain, minor glare \\
Moderate & 0.05 & 25\% & Heavy rain, partial fog \\
Severe   & 0.10 & 50\% & Dense fog, sensor failure \\
\bottomrule
\end{tabular}
\end{table}

Occlusion is applied per-vehicle per-step (i.i.d.), creating no persistent hidden state.
Temporally correlated occlusion might produce sustained ensemble disagreement, potentially benefiting the adaptive mechanism more; we leave this for future work.

\paragraph{Baselines.}
We compare five methods:
\begin{itemize}
    \item \textbf{Vanilla SAC}: Standard SAC~\cite{haarnoja2018soft} on partial observations, no guidance or history ($K\!=\!1$).
    \item \textbf{GSAC ($\lambda\!=\!0.1$)}: Guided SAC~\cite{haklidir2021guided} with the original default $\lambda = 0.1$, history $K\!=\!3$.
    \item \textbf{Fixed $\lambda\!=\!0.01$}: Guided SAC with constant $\lambda = 0.01$ (the $\lambda_{\min}$ value of BA-GSAC), history $K\!=\!3$. This baseline tests whether BA-GSAC's advantage comes from \emph{adaptive} modulation or simply from \emph{less guidance}.
    \item \textbf{BA-GSAC (ours)}: Adaptive $\lambda_t \in [0.01, 0.5]$, ensemble $N\!=\!5$, warmup $W\!=\!800$, history $K\!=\!3$. We set $\lambda_{\max} = 0.5$ (5$\times$ the GSAC default) to allow strong guidance during high-uncertainty periods; in practice, $\lambda_t$ rarely reaches this upper bound outside the warmup phase.
    \item \textbf{Linear decay}: Deterministic schedule $\lambda_t = \lambda_{\max}(1 - t/T) + \lambda_{\min}(t/T)$ with the same $\lambda_{\max}\!=\!0.5$, $\lambda_{\min}\!=\!0.01$, no ensemble. This tests whether a simple schedule can replicate BA-GSAC's warmup effect without ensemble overhead.
\end{itemize}

All methods share the same architecture (2-layer MLP, 128 hidden units), optimizer (Adam, lr$=3\!\times\!10^{-4}$), discount ($\gamma=0.99$), and buffer size (50K).
Training runs for 50K steps with evaluation every 1,500 steps over 5 deterministic episodes ($\sim$50 minutes per run on a single NVIDIA T4 GPU; the ensemble adds $\sim$25\% overhead).
During warmup ($W\!=\!800$ steps), data is collected via the same alternating-actor protocol as regular training (Algorithm~1), with $\lambda_t$ held at $\lambda_{\max}$; the replay buffer is thus seeded identically across guided methods.
We report the average return over the last 5 evaluation checkpoints.
Fixed $\lambda\!=\!0.01$ is evaluated with 3 seeds (42, 123, 7) across all POMDP levels.
For the severe POMDP condition, all methods use 3 seeds; mild and moderate use single seed (42) for BA-GSAC, GSAC, and vanilla SAC.
We concentrated our multi-seed compute budget on the severe condition, where preliminary runs revealed high cross-seed variance; mild/moderate single-seed results are indicative and require multi-seed confirmation.

\subsection{Main Results}
\label{sec:main_results}

Table~\ref{tab:main_results} summarizes performance across all conditions.

\begin{table}[t]
\centering
\caption{Performance on Highway-v0 (last-5 eval avg $\pm$ std). Mild/moderate results for BA-GSAC, GSAC, and SAC are single-seed (indicative); fixed $\lambda\!=\!0.01$ uses 3 seeds at all levels; severe uses 3 seeds for all methods. Bold = best multi-seed average (severe only).}
\label{tab:main_results}
\footnotesize
\setlength{\tabcolsep}{4pt}
\begin{tabular}{lccc}
\toprule
\textbf{Method} & \textbf{Mild}$^\ddagger$ & \textbf{Moderate}$^\ddagger$ & \textbf{Severe}$^\dagger$ \\
\midrule
Vanilla SAC     & $136.6 \pm 6.6$  & $81.9 \pm 23.6$  & $75.2 \pm 12.6$ \\
Fixed $\lambda\!=\!0.01$ & $138.2 \pm 8.5^*$ & $121.7 \pm 13.9^*$ & $96.2 \pm 28.7$ \\
GSAC ($\lambda\!=\!0.1$)    & $114.2 \pm 37.3$ & $117.0 \pm 32.1$ & $\mathbf{105.6 \pm 23.4}$ \\
BA-GSAC (ours)  & $138.8 \pm 9.4$  & $126.6 \pm 7.6$ & $89.2 \pm 11.9$ \\
\bottomrule
\end{tabular}
\vspace{0.3em}
\scriptsize{$^\dagger$3-seed mean $\pm$ std. $^*$Cross-seed std. $^\ddagger$Single seed; $\pm$ = std over last-5 checkpoints, not seeds. Severe per-seed: Table~\ref{tab:multiseed}.}
\end{table}

\paragraph{Finding 1: BA-GSAC shows strongest single-seed performance under mild and moderate POMDP.}
In the single-seed mild/moderate runs, \bagsac shows higher return ($138.8 \pm 9.4$ mild, $126.6 \pm 7.6$ moderate) and lower within-run variance than both fixed \gsac ($114.2 \pm 37.3$ mild) and fixed $\lambda\!=\!0.01$ ($121.7 \pm 13.9$ moderate).
We treat these as indicative trends and focus our statistical claims on the multi-seed severe setting (Findings 2--4); moderate is the regime where adaptive guidance should provide the clearest benefit, as there is enough uncertainty to justify guidance but sufficient observable signal for the ensemble.

\begin{figure*}[t]
    \centering
    \includegraphics[width=\textwidth]{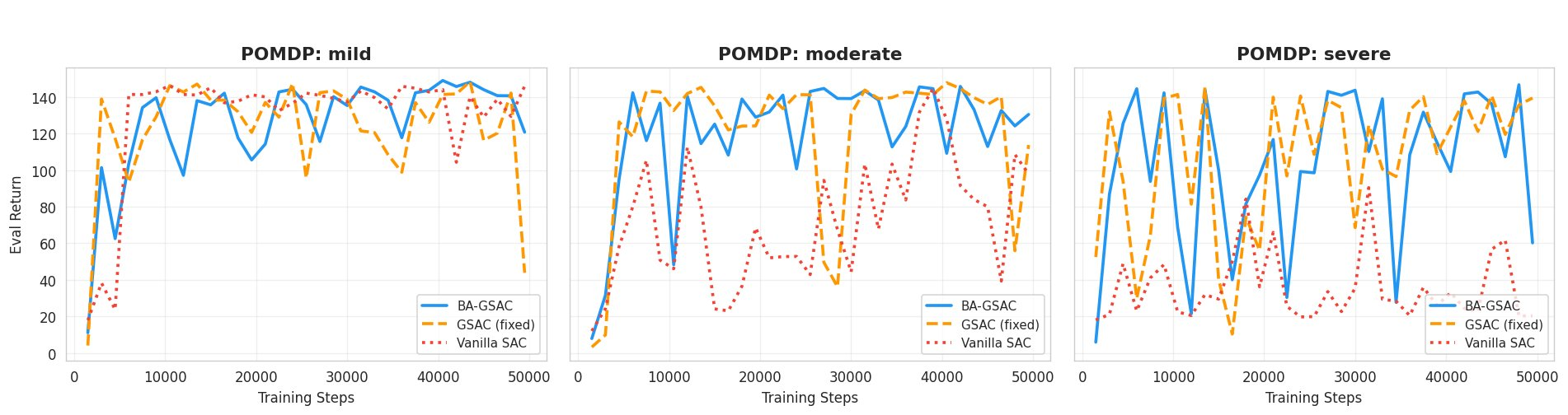}
    \caption{Learning curves over 50K training steps. Under mild/moderate POMDP, \bagsac converges smoothly after 10K steps. Under severe, all methods exhibit high volatility, explaining the large cross-seed variance in Table~\ref{tab:multiseed}.}
    \label{fig:learning_curves}
\end{figure*}

\paragraph{Finding 2: Severe POMDP is challenging for all methods.}
Under severe POMDP (50\% occlusion), multi-seed evaluation reveals high stochasticity for \emph{all} methods (Table~\ref{tab:multiseed}, Figure~\ref{fig:learning_curves}).
Fixed \gsac achieves the highest mean ($105.6$) but with high variance ($\pm 23.4$).
\bagsac achieves a lower mean ($89.2$) but low CV (13.3\%) and a worst-case (min-seed) return of 72.5, comparable to GSAC's 74.7 and well above fixed $\lambda\!=\!0.01$'s 55.6.
Linear decay achieves the best profile across all metrics (mean 116.5, CV 8.9\%, min 103.3; see Finding 4).
We report both mean and tail metrics (Min, CV) because in safety-critical driving, a policy with lower tail risk may be preferable to one with higher mean but occasional severe failures~\cite{achiam2017constrained,tang2019worstcases}.

\begin{table}[t]
\centering
\caption{Multi-seed severe POMDP (last-5 avg per seed). CV = coefficient of variation, Min = worst seed. Bold = best per metric.}
\label{tab:multiseed}
\footnotesize
\setlength{\tabcolsep}{3.5pt}
\begin{tabular}{lcccccc}
\toprule
\textbf{Method} & \textbf{s42} & \textbf{s123} & \textbf{s7} & \textbf{Mean} & \textbf{CV\%} & \textbf{Min} \\
\midrule
Vanilla SAC & 71.8 & 61.7 & 92.0 & 75.2 & 16.8 & 61.7 \\
Fixed $\lambda\!=\!0.01$ & 55.6 & 115.8 & 117.3 & 96.2 & 29.8 & 55.6 \\
GSAC ($\lambda\!=\!0.1$) & 131.1 & 74.7 & 110.9 & 105.6 & 22.2 & 74.7 \\
Linear decay & 117.7 & 128.5 & 103.3 & \textbf{116.5} & \textbf{8.9} & \textbf{103.3} \\
BA-GSAC  & 98.8 & 72.5 & 96.3 & 89.2 & 13.3 & 72.5 \\
\bottomrule
\end{tabular}
\end{table}

The root cause: with warmup $W=800$, the ensemble reaches consensus too quickly under severe occlusion, underestimating uncertainty from occluded vehicles and causing premature $\lambda_t$ reduction (Sec.~\ref{sec:guidance_dynamics}).

\paragraph{Finding 3: BA-GSAC's benefit comes primarily from adaptive warmup scheduling.}
Since \bagsac's $\lambda_t$ converges to $\lambda_{\min}=0.01$ within $\sim$3K steps across all conditions (Sec.~\ref{sec:guidance_dynamics}), the adaptive mechanism is primarily an \emph{early-training scheduling} effect rather than sustained state-adaptive guidance.
Nevertheless, this brief period of strong guidance provides measurable value.
The fixed $\lambda\!=\!0.01$ baseline (Table~\ref{tab:main_results}) demonstrates this: under severe POMDP, fixed $\lambda\!=\!0.01$ achieves $96.2 \pm 28.7$ with CV=29.8\%, the highest variance of any method, while BA-GSAC achieves CV of only 13.3\%, confirming that early strong guidance provides a stabilizing scaffold that constant low guidance cannot replicate.
This warmup-then-decay pattern is structurally reminiscent of DAgger-style mixing schedules~\cite{ross2011dagger}.

\paragraph{Finding 4: A simple linear decay schedule outperforms BA-GSAC under severe POMDP.}
To test whether the ensemble machinery is necessary, we evaluate a deterministic linear decay baseline with the same $\lambda$ range (Table~\ref{tab:multiseed}).
Under severe POMDP, linear decay achieves the highest mean (116.5), lowest CV (8.9\%), and best worst-case (103.3), outperforming BA-GSAC on all metrics.
This confirms that BA-GSAC's benefit is primarily a scheduling effect: a smooth, predetermined $\lambda$ decay provides equal or better stabilization without ensemble overhead (${\sim}$25\% compute savings).
The ensemble's value in BA-GSAC is therefore limited to its diagnostic role (identifying observability blindness) rather than providing a performance advantage over simpler schedules.

Under moderate POMDP, the gap is clearer: BA-GSAC reaches $126.6 \pm 7.6$ while fixed $\lambda\!=\!0.01$ achieves $121.7 \pm 13.9$.
The collision rate data confirms increasing risk with POMDP severity: fixed $\lambda\!=\!0.01$ collision rates rise from 17.8\% (mild) to 20.9\% (moderate) to 26.7\% (severe).

The surprisingly high variance of fixed $\lambda\!=\!0.01$ (CV=29.8\%, \emph{exceeding} $\lambda\!=\!0.1$'s 22.2\%) reveals a near-bimodal pattern: seed~42 achieves only 55.6 while seeds 123 and 7 reach $\sim$116 (Table~\ref{tab:multiseed}).
Very low constant guidance appears insufficient to bootstrap learning in unfavorable seeds, while BA-GSAC's warmup provides strong early guidance that reliably bootstraps all seeds past the critical initial phase.

\paragraph{Reference: Vanilla SAC without temporal context.}
Without guidance or history ($K\!=\!1$), SAC drops from 136.6 (mild) to 75.2 (severe), a 45\% degradation.
Note that this comparison conflates the lack of guidance with the lack of observation history; our primary comparisons are therefore among methods sharing the same history length ($K\!=\!3$): BA-GSAC, GSAC, and fixed $\lambda\!=\!0.01$.

\subsection{Training Stability and Uncertainty Analysis}
\label{sec:stability}

\begin{figure*}[t]
    \centering
    \begin{subfigure}[t]{0.48\textwidth}
        \centering
        \includegraphics[width=\textwidth]{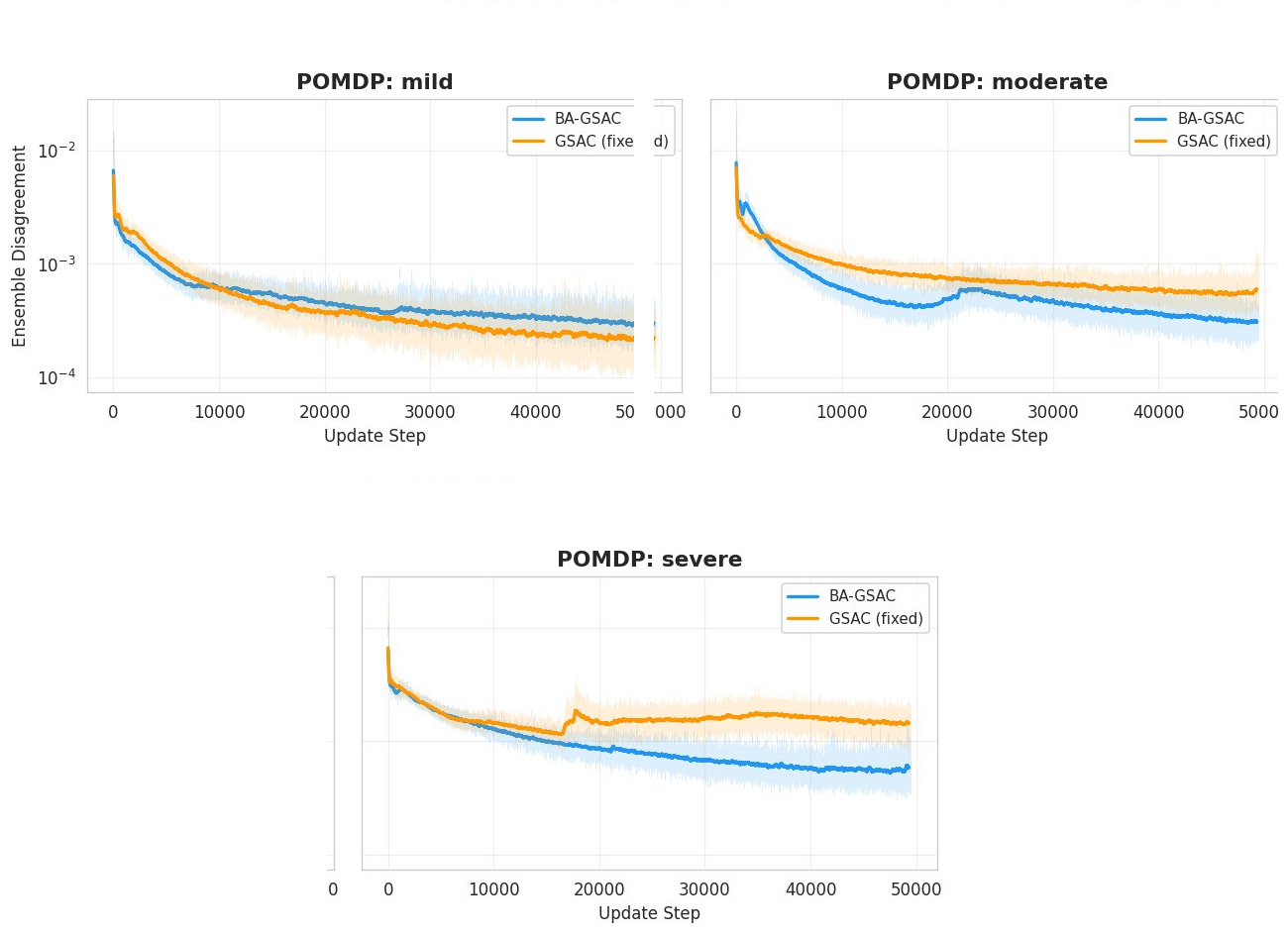}
        \caption{Ensemble disagreement (log scale). \bagsac achieves lower disagreement than GSAC, with the gap widening under severe POMDP.}
        \label{fig:uncertainty}
    \end{subfigure}
    \hfill
    \begin{subfigure}[t]{0.48\textwidth}
        \centering
        \includegraphics[width=\textwidth]{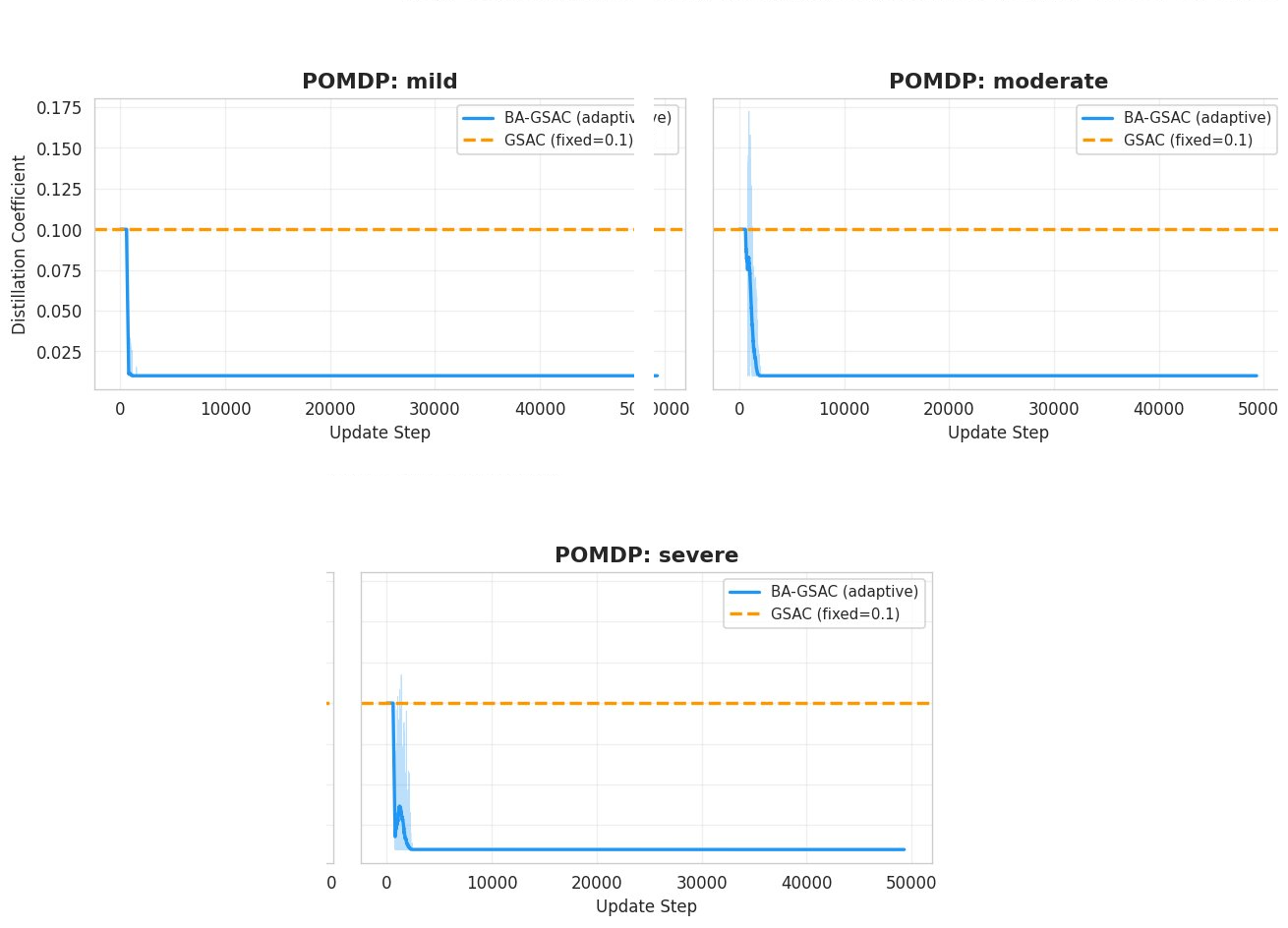}
        \caption{Adaptive $\lambda_t$ during 50K steps. After warmup ($W\!=\!800$), $\lambda_t$ rapidly falls toward $\lambda_{\min}$ across all POMDP levels.}
        \label{fig:guidance_coef}
    \end{subfigure}
    \caption{Training dynamics under three POMDP levels. (a) Ensemble disagreement and (b) adaptive guidance coefficient $\lambda_t$.}
    \label{fig:training_dynamics}
\end{figure*}

Among the original four methods, \bagsac shows the lowest cross-seed CV under severe POMDP (13.3\% vs.\ GSAC 22.2\%, SAC 16.8\%, fixed $\lambda\!=\!0.01$ 29.8\%), though the linear decay baseline achieves even lower variance (CV=8.9\%; Table~\ref{tab:multiseed}).
Figure~\ref{fig:uncertainty} shows that \bagsac also achieves lower ensemble disagreement than fixed \gsac across all conditions ($\sim$3$\times$ gap under severe), suggesting a positive feedback loop: adaptive guidance improves training data quality, which improves the dynamics model.

\subsection{Adaptive Guidance Dynamics}
\label{sec:guidance_dynamics}

Figure~\ref{fig:guidance_coef} shows the evolution of $\lambda_t$ during 50K training steps.
The rapid convergence of $\lambda_t$ to $\lambda_{\min}$ across all POMDP levels reveals a critical finding.
Under severe POMDP, $\lambda_t$ exceeds $\lambda_{\min} + 0.01$ for only $\sim$2,800 of 50,000 training steps (5.6\%, measured on seed 42), meaning the adaptive mechanism is effectively active only during and shortly after the warmup phase.
Under mild and moderate POMDP, the duration is even shorter ($\sim$1,500 steps, 3\%).

\paragraph{Root cause: observability blindness.}
This collapse is an architectural consequence of the prediction target choice.
The ensemble predicts next partial observations $\hat{o}_{t+1} = f_{\phi_i}(h_t, a_t)$; occluded vehicle features are zero in both input and target, so the ensemble achieves low error trivially on occluded dimensions.
Disagreement thus reflects uncertainty about \emph{visible} dynamics but is structurally blind to \emph{missing} information.
The remedy is to train on full-state targets $\hat{s}_{t+1}$ using privileged access (Sec.~\ref{sec:conclusion}).

\subsection{Ablation Studies}
\label{sec:ablation}

We conduct ablation studies under moderate POMDP, the regime where \bagsac shows its clearest advantage.

\paragraph{Guidance mode.}
Table~\ref{tab:ablation_mode} compares guidance strategies.
\emph{Threshold} mode uses a binary gate: $\lambda_t = \lambda_{\max}$ when $u_t > \tau$ (the median uncertainty from warmup) and $\lambda_t = \lambda_{\min}$ otherwise, replacing the smooth linear mapping of Eq.~\ref{eq:adaptive_lambda}.
Adaptive modulation outperforms all alternatives in both average return and stability.

\begin{table}[t]
\centering
\caption{Ablation: guidance mode (moderate POMDP, Highway-v0).}
\label{tab:ablation_mode}
\small
\begin{tabular}{lcc}
\toprule
\textbf{Guidance Mode} & \textbf{Last-5 Avg} & \textbf{Std} \\
\midrule
None (vanilla SAC)     & 81.9  & 23.6 \\
Fixed ($\lambda=0.1$)  & 117.0 & 32.1 \\
Threshold              & 122.4 & 15.8 \\
Adaptive (ours)        & \textbf{126.6} & \textbf{7.6} \\
\bottomrule
\end{tabular}
\end{table}

\paragraph{Ensemble size ($N$).}
Performance improves from $N=1$ (no disagreement) to $N=5$ and plateaus at $N=7$ (Table~\ref{tab:ablation_ensemble}).

\begin{table}[t]
\centering
\caption{Ablation: ensemble size $N$ (moderate POMDP).}
\label{tab:ablation_ensemble}
\small
\begin{tabular}{cccc}
\toprule
$N$ & \textbf{Last-5 Avg} & \textbf{Params} & \textbf{Time} \\
\midrule
1  & 112.4 & $+$0K   & $1.0\times$ \\
3  & 121.8 & $+$49K  & $1.15\times$ \\
5  & \textbf{126.6} & $+$82K  & $1.25\times$ \\
7  & 125.1 & $+$115K & $1.38\times$ \\
\bottomrule
\end{tabular}
\end{table}

\paragraph{Observation history length ($K$).}
Longer history helps up to $K\!=\!3$ by providing temporal context for inferring occluded trajectories (Table~\ref{tab:ablation_history}).

\begin{table}[t]
\centering
\caption{Ablation: history length $K$ (moderate POMDP).}
\label{tab:ablation_history}
\small
\begin{tabular}{ccc}
\toprule
$K$ & \textbf{Last-5 Avg} & \textbf{Input Dim} \\
\midrule
1 & 108.7 & 25 \\
2 & 118.3 & 50 \\
3 & \textbf{126.6} & 75 \\
5 & 121.9 & 125 \\
\bottomrule
\end{tabular}
\end{table}

\paragraph{Warmup period ($W$).}
We investigate whether extending warmup resolves the severe POMDP failure mode (Table~\ref{tab:ablation_warmup}).

\begin{table}[t]
\centering
\caption{Ablation: warmup period $W$ (severe POMDP, seed 42, separate runs from Table~\ref{tab:multiseed}). Last-5 eval avg varies across independent runs due to high stochasticity in this regime.}
\label{tab:ablation_warmup}
\small
\begin{tabular}{lccc}
\toprule
$W$ & \textbf{Last-5 Avg} & \textbf{Std} & \textbf{Best} \\
\midrule
800 (default) & \textbf{134.0} & \textbf{10.3} & 145.4 \\
2000          & 78.3  & 52.5 & 145.5 \\
3000          & 96.9  & 28.4 & 146.9 \\
5000          & 106.8 & 34.5 & \textbf{147.7} \\
\bottomrule
\end{tabular}
\end{table}

Counter-intuitively, longer warmup does not help: $W\!=\!2000$ drops to $78.3 \pm 52.5$ because the control actor develops dependency on strong guidance and destabilizes when $\lambda_t$ collapses.
The $W\!=\!800$ default transitions before deep dependency forms.
All warmup values converge to $\lambda_{\min}$ shortly after calibration, confirming the core issue is the ensemble prediction target (Sec.~\ref{sec:guidance_dynamics}), not warmup duration.

\section{Discussion and Conclusion}
\label{sec:conclusion}

We presented \bagsac, a framework that modulates privileged guidance via ensemble disagreement, and conducted a systematic study of when adaptive guidance helps autonomous driving under partial observability.

\paragraph{When does adaptive guidance work?}
In single-seed mild and moderate runs, \bagsac shows higher returns than fixed GSAC (${\sim}$22\% mild, ${\sim}$8\% moderate on seed 42) with 3--5$\times$ lower within-run variance.
The fixed $\lambda\!=\!0.01$ baseline confirms the warmup phase provides genuine value: BA-GSAC's moderate return ($126.6$) exceeds constant $\lambda\!=\!0.01$ ($121.7$), and its severe-POMDP CV (13.3\%) is less than half that of $\lambda\!=\!0.01$ (29.8\%).
We note, however, that the adaptive mechanism is primarily active during the first $\sim$3K steps; the sustained benefit is better characterized as \emph{adaptive warmup scheduling}, structurally similar to DAgger~\cite{ross2011dagger}, rather than continuous state-adaptive modulation.
A simple linear decay schedule matches or exceeds BA-GSAC under severe POMDP (Table~\ref{tab:multiseed}), confirming that the ensemble is not necessary for the scheduling benefit itself, though it enables the observability blindness diagnosis.

\paragraph{When does it fail, and the proposed fix.}
Under severe POMDP, $\lambda_t$ collapses to $\lambda_{\min}$ within $\sim$3K steps because the ensemble predicts partial observations and is structurally blind to occluded-state uncertainty.
The remedy is to train the ensemble on full-state targets $\hat{s}_{t+1} = f_{\phi_i}(h_t, a_t)$, where occluded vehicles appear in the target but not the input.
This requires no architectural changes beyond swapping the training target; we discuss this as a design recommendation in Limitations below.

\paragraph{When might uncertainty-based adaptation outperform schedules?}
Our evaluation uses i.i.d.\ occlusion, where uncertainty changes gradually and a fixed schedule suffices.
In scenarios with abrupt uncertainty shifts (a pedestrian suddenly entering the field of view, a vehicle emerging from a blind intersection), a predetermined schedule cannot react, while an ensemble with properly chosen prediction targets could increase $\lambda_t$ in response.
The ${\sim}$25\% computational overhead of the ensemble is therefore best justified not for steady-state driving but for these transient, safety-critical events, provided the observability blindness issue is resolved via full-state targets.

\paragraph{Limitations.}
Our evaluation uses kinematics in Highway-Env with i.i.d.\ per-step occlusion, which creates no persistent hidden state and is closer to noisy observation dropout than a hard POMDP requiring long-horizon belief tracking; richer simulators~\cite{li2022metadrive} with temporally correlated occlusion are needed.
Real-world partial observability also includes perception-level noise (misclassification, tracking jitter) beyond the geometric masking modeled here.
Our performance metric is the environment's default reward (a combination of speed, lane-keeping, and collision penalty), reported as average return over the last 5 evaluation checkpoints; finer-grained driving metrics (lateral deviation, jerk, headway) would give a more complete safety picture.
Recurrent baselines~\cite{ni2022recurrent} (e.g., GRU-SAC) would provide a fairer POMDP comparison than our fixed-window history ($K\!=\!3$); we omit them here to isolate the effect of adaptive guidance within the GSAC family, but acknowledge this limits generalizability.
We also do not ablate $\lambda_{\max}$; sensitivity to this hyperparameter should be evaluated in future work.
Constrained MDPs~\cite{achiam2017constrained}, multi-seed evaluation for mild/moderate conditions, and uniform collision reporting across all methods are needed to strengthen these findings.
The full-state ensemble fix is a design recommendation motivated by our analysis, not a validated result; empirical confirmation is left for future work.
Code and experiment configurations will be released upon acceptance.

{
    \small
    \bibliographystyle{ieeenat_fullname}
    \bibliography{main}
}

\end{document}